\documentclass[11pt,a4paper]{article}
\usepackage[hyperref]{acl2021}
\usepackage{times}
\usepackage{latexsym}

\usepackage{microtype}
\usepackage{graphicx}
\usepackage{multirow}
\usepackage{multicol}
\usepackage{amsmath}
\usepackage{cases}
\usepackage{array}
\usepackage{algorithm}
\usepackage{algpseudocode}

\aclfinalcopy 

\title{Coreference-Aware Dialogue Summarization}

\author{Zhengyuan Liu, \ Ke Shi, \ Nancy F. Chen \\
  Institute for Infocomm Research, A*STAR, Singapore \\
  \texttt{\{liu\_zhengyuan,shi\_ke,nfychen\}@i2r.a-star.edu.sg}}

\begin{document}
\maketitle
\begin{abstract}
Summarizing conversations via neural approaches has been gaining research traction lately, yet it is still challenging to obtain practical solutions. Examples of such challenges include unstructured information exchange in dialogues, informal interactions between speakers, and dynamic role changes of speakers as the dialogue evolves. Many of such challenges result in complex coreference links. Therefore, in this work, we investigate different approaches to explicitly incorporate coreference information in neural abstractive dialogue summarization models to tackle the aforementioned challenges. 
Experimental results show that the proposed approaches achieve state-of-the-art performance, implying it is useful to utilize coreference information in dialogue summarization. Evaluation results on factual correctness suggest such coreference-aware models are better at tracing the information flow among interlocutors and associating accurate status/actions with the corresponding interlocutors and person mentions.
\end{abstract}

\section{Introduction}
\label{introduction}
\begin{figure}[ht]
    \centering
    \includegraphics[width=0.46\textwidth]{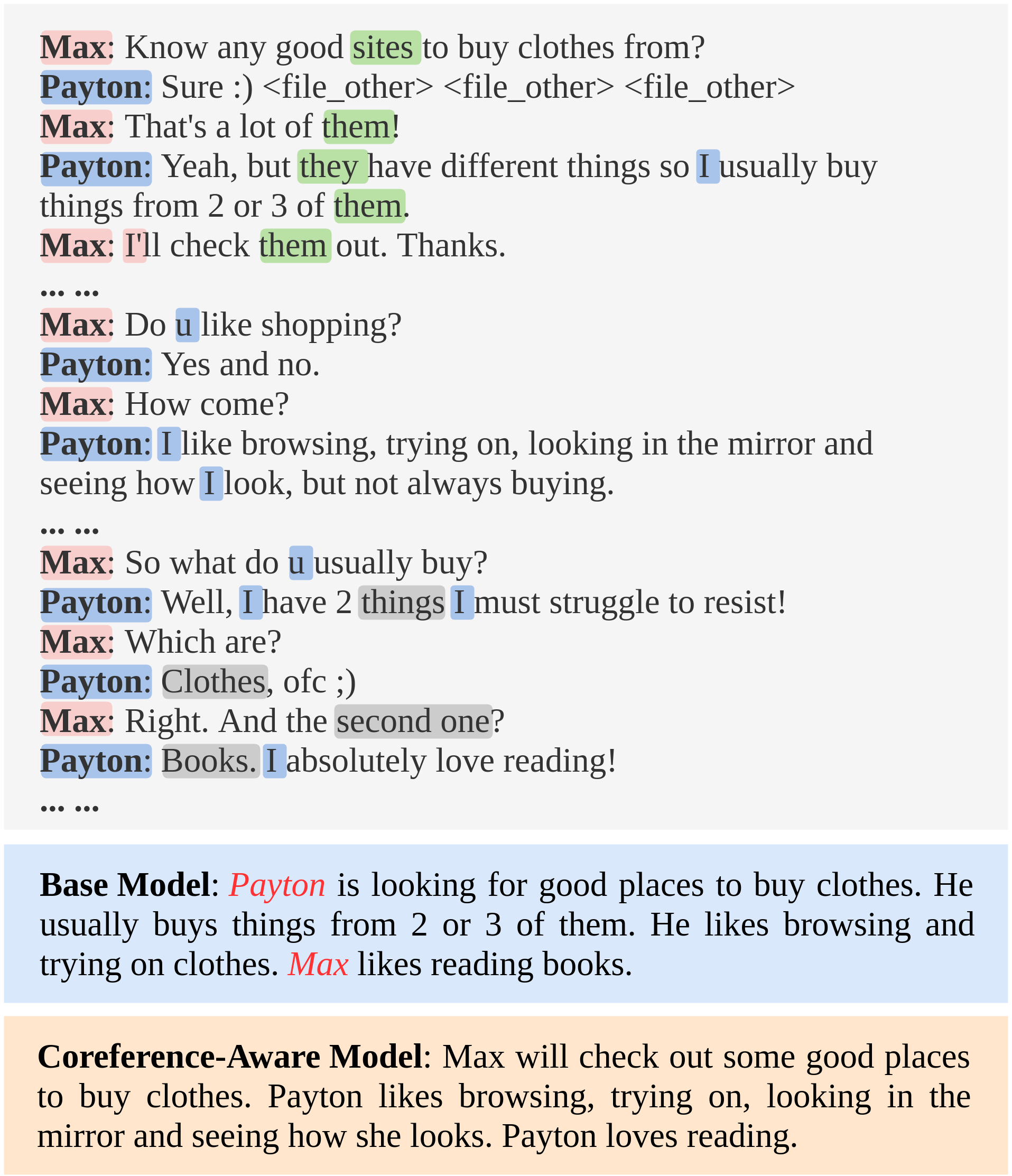}
    \caption{An example of dialogue summarization: The original conversation (in grey) is abbreviated; the summary generated by a baseline model is in blue; the summary generated by a coreference-aware model is in orange. While these two summaries obtain similar ROUGE scores, the summary from the baseline model is not factually correct; errors are highlighted in italic and magenta.}
    \label{fig:example}
\vspace{-0.4cm}
\end{figure}

Text summarization condenses the source content into a shorter version while retaining essential and informative content. Most prior work focuses on summarizing well-organized single-speaker content such as news articles \cite{hermann2015teaching} and encyclopedia documents \cite{liu2018wikiSumm}. Recently, models applied on text summarization benefit favorably from sophisticated neural architectures and pre-trained contextualized language backbones: on the popular benchmark corpus CNN/Daily Mail \cite{hermann2015teaching}, \citet{liu-lapata-2019-text} explored fine-tuning BERT \cite{devlin-etal-2019-bert} to achieve state-of-the-art performance for extractive news summarization, and BART \cite{lewis-etal-2020-bart} has also improved generation quality on abstractive summarization.

While there has been substantial progress on document summarization, dialogue summarization has received less attention. Unlike documents, conversations are interactions among multiple speakers, they are less structured and are interspersed with more informal linguistic usage \citep{SACKS19787}.
Based on the characteristics of human-to-human conversations \cite{jurafsky2008speech}, challenges of summarizing dialogues stem from: (1) Multiple speakers:
the interactive information exchange among interlocutors implies that essential information is referred to back and forth across speakers and dialogue turns; 
(2) Speaker role shifting: multi-turn dialogues often involve frequent role shifting from one type of interlocutor to another type (e.g., questioner becomes responder and vice versa); (3) Ubiquitous referring expressions: aside from speakers referring to themselves and each other, speakers also mention third-party persons, concepts, and objects. Moreover, referring could also take on forms such as anaphora or cataphora where pronouns are used, making coreference chains more elusive to track. 
Figure \ref{fig:example} shows one dialogue example: two speakers exchange information among interactive turns, where the pronoun \textit{``them''} is used multiple times, referring to the word \textit{``sites''}. Without sufficient understanding of the coreference information, the base summarizer fails to link mentions with their antecedents, and produces an incorrect description (highlighted in magenta and italic) in the generation.
From the aforementioned linguistic characteristics, dialogues possess multiple inherent sources of complex coreference, motivating us to explicitly consider coreference information for dialogue summarization to more appropriately model the context, to more dynamically track the interactive information flow throughout a conversation, and to enable the potential of multi-hop dialogue reasoning.

Previous work on dialogue summarization focuses on  modeling conversation topics or dialogue acts \citep{Goo2018Abs, liu2019topic, li-etal-2019-keep, chen-yang-2020-multi}. Few, if any, leverage on features from coreference information explicitly. On the other hand, large-scale pre-trained language models are shown only to implicitly model lower-level linguistic knowledge such as part-of-speech and syntactic structure \citep{tenney2018what, jawahar-etal-2019-bert}.
Without directly training on tasks that provide specific and explicit linguistic annotation such as coreference resolution or semantics-related reasoning, model performance remains subpar for language generation tasks \citep{dasigi-etal-2019-quoref}.
Therefore, in this paper, we propose to improve abstractive dialogue summarization by explicitly incorporating coreference information.
Since entities are linked to each other in coreference chains, we postulate adding a graph neural layer could readily characterize the underlying structure, thus enhancing contextualized representation. We further explore two parameter-efficient approaches: one with an additional coreference-guided attention layer, and the other resourcefully enhancing BART’s limited coreference resolution capabilities by conducting probing analysis to augment our coreference injection design.

Experiments on SAMSum \citep{gliwa-etal-2019-samsum} show that the proposed methods achieve state-of-the-art performance. Furthermore, human evaluation and error analysis suggest our models generate more factually consistent summaries. As shown in Figure \ref{fig:example}, a model guided with coreference information accurately associates events with their corresponding subjects, and generates more trustworthy summaries compared with the baseline.

\section{Related Work}
In abstractive text summarization, recent studies mainly focus on neural approaches. \citet{rush-etal-2015-neural} proposed an attention-based neural summarizer with sequence-to-sequence generation. Pointer-generator networks \cite{see2017get} were designed to directly copy words from the source content, which resolved out-of-vocabulary issues. \citet{liu-lapata-2019-text} leveraged the pre-trained language model BERT \cite{devlin-etal-2019-bert} on both extractive and abstractive summarization. \citet{lewis-etal-2020-bart} proposed BART, taking advantage of the bi-directional encoder in BERT and the auto-regressive decoder of GPT \citep{radford2018improving} to obtain impressive results on language generation.

While many prior studies focus on summarizing well-organized text such as news articles \citep{hermann2015teaching}, dialogue summarization has been gaining traction.  \citet{shang-etal-2018-unsupervised} proposed an unsupervised multi-sentence compression method for meeting summarization. \citet{Goo2018Abs} introduced a sentence-gated mechanism to grasp the relations between dialogue acts. \citet{liu2019topic} proposed to utilize topic segmentation and turn-level information \cite{liu2019readingTurn} for conversational tasks. \citet{zhao2019abstractive} proposed a neural model with a hierarchical encoder and a reinforced decoder to generate meeting summaries. \citet{chen-yang-2020-multi} used diverse conversational structures like topic segments and conversational stages to design a multi-view summarizer, and achieved the current state-of-the-art performance on the SAMSum corpus \citep{gliwa-etal-2019-samsum}.

\begin{figure*}[t!]
    \begin{center}
    \includegraphics[width=0.92\textwidth]{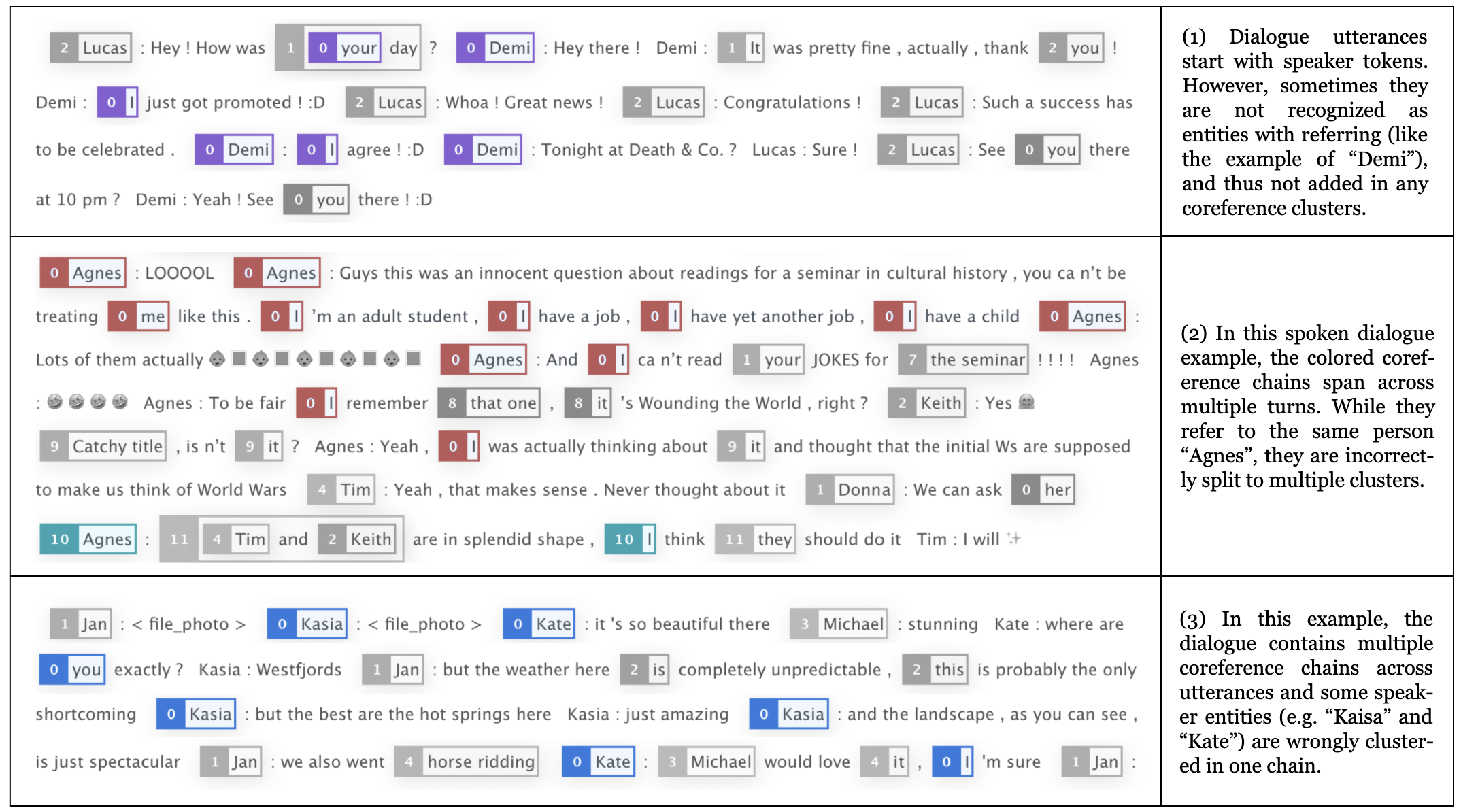}
    \end{center}
    \caption{Examples of three common issues in adopting a document coreference resolution model for dialogues without additional domain adaptation training. Spans in blocks are items in coreference clusters with their cluster ID number. We highlight some spans for better readability.}
    \label{fig:coref_big_example}
    \vspace{-0.3cm}
\end{figure*}

Improving factual correctness has received keen attention in neural abstractive summarization lately. \citet{cao2018faithful} leveraged on dependency parsing and open information extraction to enhance the reliability of generated summaries. \citet{zhu2020boosting} proposed a factual corrector model based on knowledge graphs, significantly improving factual correctness in text summarization.

\section{Dialogue Coreference Resolution}
\label{sec:coref_resolution}
Since the common summarization datasets do not contain coreference annotations, automatic coreference resolution is needed to process the samples. Neural approaches \cite{joshi-etal-2020-spanbert} have shown impressive performance on document coreference resolution. However, they are still sub-optimal for conversational scenarios \cite{chen2017robust}, and there are no large-scale annotated dialogue corpora for transfer learning. When applying a document coreference resolution model \cite{Lee2017EndtoendNC, joshi-etal-2020-spanbert} on dialogue samples without domain adaptation,\footnote{The off-the-shelf version of coreference resolution model we used is \textit{allennlp-public-models/coref-spanbert-large-2021.03.10}, which is trained on OntoNotes 5.0 dataset.} as shown in Figure \ref{fig:coref_big_example}, we observed some common issues:
(1) Each dialogue utterance starts with a speaker, but sometimes this speaker is not recognized as a coreference-related entity, and thus not added in any coreference clusters;
(2) In dialogues, coreference chains are often spanned across multiple turns, but sometimes they are split to multiple clusters;
(3) When a dialogue contains multiple coreference chain across multi-turns, speaker entities could be wrongly clustered.

Based on the observation, to improve the overall quality of dialogue coreference resolution, we conducted data post-processing on the automatic output: (1) First, we applied a model ensemble strategy to obtain more accurate cluster predictions; (2) Then, we re-assigned coreference cluster labels to the words with speaker roles that were not included in any chains; (3) Moreover, we compared the clusters and merged those that presented the same coreference chain. Human evaluation on the processed data showed that this post-processing reduced incorrect coreference assignments by approximately 19\%.\footnote{In our pilot experiment, we observed that models with original coreference resolution outputs showed 10\% relative lower performance than that with the optimized data, validating the effectiveness of our post-processing.}

\section{Coreference-Aware Summarization}
In this section, we adopt a neural model for abstractive dialogue summarization, and investigate various methods to enhance it with the coreference information obtained in Section \ref{sec:coref_resolution}.

The base neural architecture is a sequence-to-sequence model Transformer \cite{vaswani2017attention}. Given a conversation containing $n$ tokens $T=\{t_1, t_2,...,t_n\}$, a self-attention-based encoder is used to produce the contextualized hidden representations $H=\{h_1,h_2,...,h_n\}$, then an auto-regressive decoder generates the target sequence $O=\{w_1,w_2,...,w_k\}$ sequentially. Here, we use BART \citep{lewis-etal-2020-bart} as the pre-trained language backbone, and conduct fine-tuning.

\begin{figure}
    \begin{center}
    \includegraphics[width=0.45\textwidth]{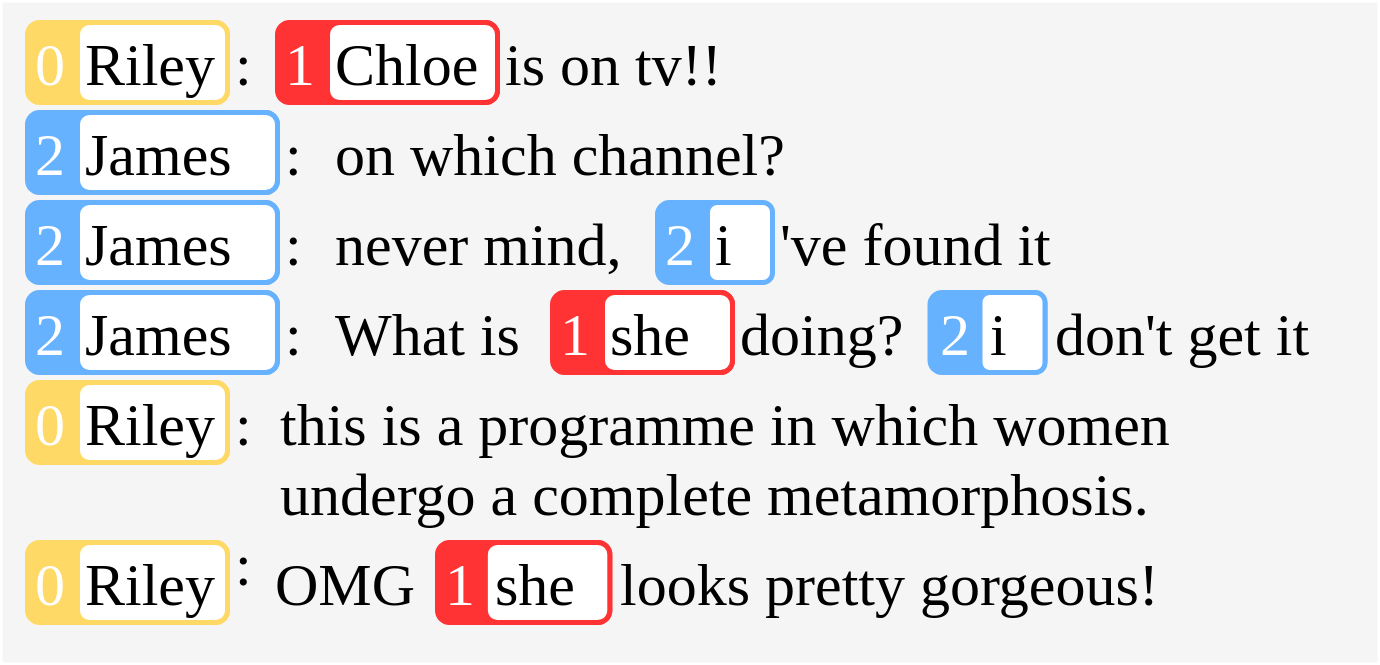}
    \end{center}
    \caption{One dialogue example with labeled coreference clusters: there are three coreference clusters in this conversation, where each cluster contains all mentions of one personal identity.}
    \label{fig:coref-example}
\vspace{-0.2cm}
\end{figure}

For each dialogue, there is a set of coreference clusters $\{C_1,C_2,...,C_u\}$, and each cluster $C_i$ contains entities $\{E^i_{1},E^i_{2}...,E^i_{m}\}$. As the multi-turn dialogue sample shown in Figure \ref{fig:coref-example}, there are three coreference clusters (colored in yellow, red, and blue, respectively), and each cluster consists a number of words/spans in the same coreference chain. 
During the conversational interaction, the referring of pronouns is important for semantic context understanding \cite{SACKS19787}, thus we postulate that incorporating coreference information explicitly can be useful for abstractive dialogue summarization. In this work, we focus on enhancing the encoder with auxiliary coreference features.

\subsection{GNN-Based Coreference Fusion}
\label{GNN}
As entities in coreference chains link to each other, a graphical representation could readily characterize the underlying structure and facilitate computational modeling of the inter-connected relations. 
In previous works, Graph Convolutional Networks (GCN) \citep{kipf2017semi} show strong capability of modeling graphical features in various tasks \cite{yasunaga-etal-2017-graph,xu-etal-2020-discourse}, thus we use it for the coreference feature fusion.

\subsubsection{Coreference Graph Construction}

To build the chain of a coreference cluster, we add links between each entity and their mentions. Unlike previous work \cite{xu-etal-2020-discourse} where entities in one cluster are all pointed to the first occurrence, here we connect the adjacent pairs to retain more local information. More specifically, given a cluster $C_i$ of entities $\{E^i_{1},E^i_{2}...,E^i_{m}\}$, we add a link of each $E$ to its precedent.

Then each coreference chain is transformed to a graph, and fed to a graph neural network (GNN). Given a text input of $n$ tokens (here we use a sub-word tokenization), a coreference graph $G$ is initialized with $n$ nodes and an empty adjacent matrix $G[:][:] = 0$. Iterating each coreference cluster $C$, the first token $t_i$ of each mention (a word or a text span) is connected with the first token $t_j$ of its antecedent in the same cluster with a bi-directional edge, \emph{i.e.,} $G[i][j] = 1$ and $G[j][i] = 1$.

\begin{figure}
    \begin{center}
    \includegraphics[width=0.48\textwidth]{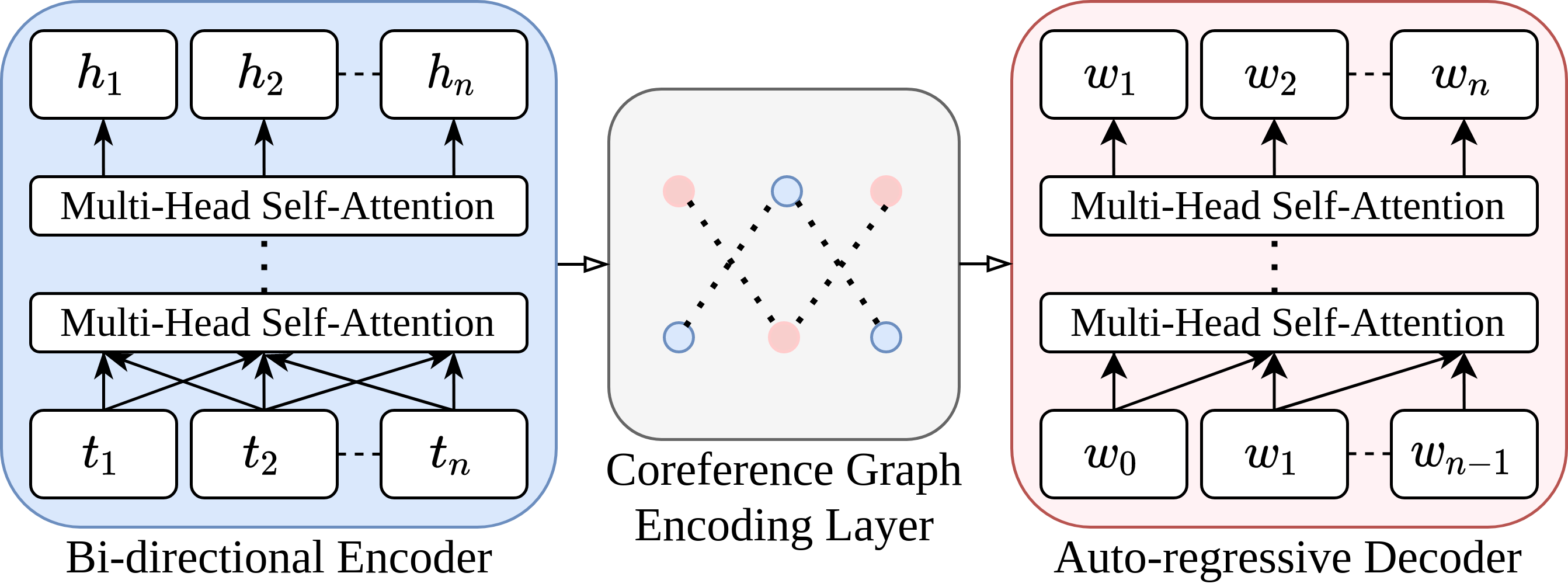}
    \end{center}
    \caption{Architecture overview of the GNN-based coreference fusion: the encoder is employed to encode the input sequence; the coreference graph encoding layer is used to model the coreference connections between all mentions; the auto-regressive decoder generates the summaries.}
    \label{fig:GNN}
\vspace{-0.2cm}
\end{figure}

\subsubsection{GNN Encoder}

Given a graph $G$ with the nodes (words/spans with coreference information in the conversation) and the edges (links between mentions), we employ stacked graph modeling layers to update the hidden representations $H$ of all nodes. Here, we take a single coreference graph encoding (CGE) layer as an example: the input of the first CGE layer is the output $H$ from the Transformer encoder. We denote the input of $k$-th CGE layer as $H^k = \{h_1^k,...,h_n^k\}$, and the representations of ($k$+1)-th layer $H^{k+1}$ are updated as follows:

\begin{equation}
\small
     u_i^k = W_1^k\mathrm{ReLU}(W_0^kh_i^k+b_0^k)+b_1^k
\end{equation}
\vspace{-0.2cm}
\begin{equation}
\small
    v_i^k = \mathrm{LayerNorm}(h_i^k+\mathrm{Dropout}(u_i^k))
\end{equation}
\vspace{-0.2cm}
\begin{equation}
\small
     w_i^k = \mathrm{ReLU}(\sum_{j\in N_i}\frac{1}{|N_i|}W_2^kv_j^k + b_2^k)
\end{equation}
\vspace{-0.2cm}
\begin{equation}
\small
    h_i^{k+1} = \mathrm{LayerNorm}(\mathrm{Dropout}(w_i^k) + v_i^k)
\end{equation}

\noindent where $W_i$ and $b_i$ denote the trainable parameter matrix and bias, $LayerNorm(*)$ is the layer normalization component, and $N_i$ denotes the neighborhood nodes of the $i$-th node. After feature propagation in all stacked CGE layers, we obtain the final representations by adding the coreference-aware hidden states $H^G = \{h_1^G,...,h_n^G\}$ with the contextualized hidden states $H$ (here a weight $\lambda$ is used, and initialized as 0.7), then the auto-regressive decoder is applied to generate summaries.

\begin{figure}
    \begin{center}
    \includegraphics[width=0.48\textwidth]{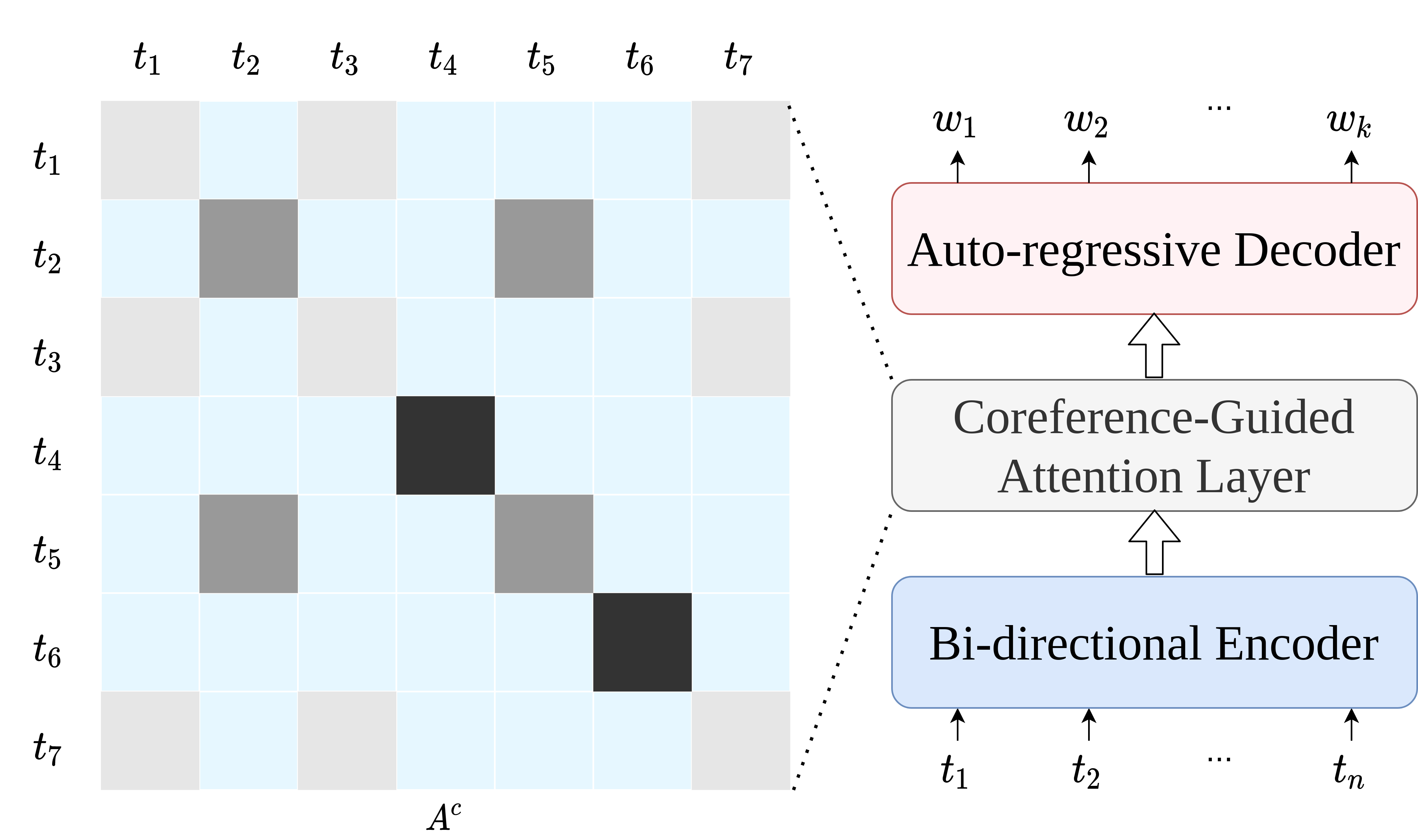}
    \end{center}
    \caption{Architecture overview of coreference-guided attention model and an example of coreference attention weight matrix $A^c$, where \{$t_1$,$t_3$,$t_7$\} are in one coreference cluster and \{$t_2$,$t_5$\} are in another cluster, while $t_4$ and $t_6$ are tokens without any coreference link.}
    \label{fig:attention}
\vspace{-0.2cm}
\end{figure}

\subsection{Coreference-Guided Attention}
\label{attention}

Aside from the GNN-based method which introduces a certain number of additional parameters, we further explore a parameter-free method. With the self-attention mechanism \cite{vaswani2017attention}, contextualized representation can be obtained with attentive weighted sum. For entities in a coreference cluster, they all share the referring information at the semantic level. Therefore, we propose to fuse the coreference information via one additional attention layer in the contextualized representation.

Given a sample with coreference clusters, a coreference-guided attention layer is constructed to update the encoded representations $H$. The overview of adding the coreference-guided attention layer is shown in Figure \ref{fig:attention}. Since items in the same coreference cluster are attended to each other, values in the attention weight matrix $A^c$ are normalized with the number of all referring mentions in one cluster, then the representation $h_i$ of token $i$ is updated according to the following:
\vspace{-0.2cm}
\begin{equation}
    a_i=\sum_{j\in C^*}\frac{1}{|C^*|}h_j, \ \ if \ t_i \in C^*
\end{equation}
\vspace{-0.2cm}
\begin{equation}
    h_i^A = \lambda h_i + (1-\lambda)a_i
\vspace{0.1cm}
\end{equation}
\noindent{where $a_i$ is the attentive representation of $t_i$, if $t_i$ belongs to one coreference cluster $C^*$, the representation of $t_i$ is updated, otherwise, it remains unchanged. $\lambda$ is an adjustable parameter and initialized as 0.7. In our experimental settings, we observed that when $\lambda$ is trainable, it is trained to be $0.69$ when our coreference-guided attention model achieved the best performance on the validation set. Following the coreference-guided attention layer, we obtain the final representations with coreference information $H^A = \{h_1^A,...,h_n^A\}$, then they are fed to the decoder for output generation.}

\subsection{Coreference-Informed Transformer}
\label{transformer}

\begin{figure}[]
    \begin{center}
    \includegraphics[width=0.43\textwidth]{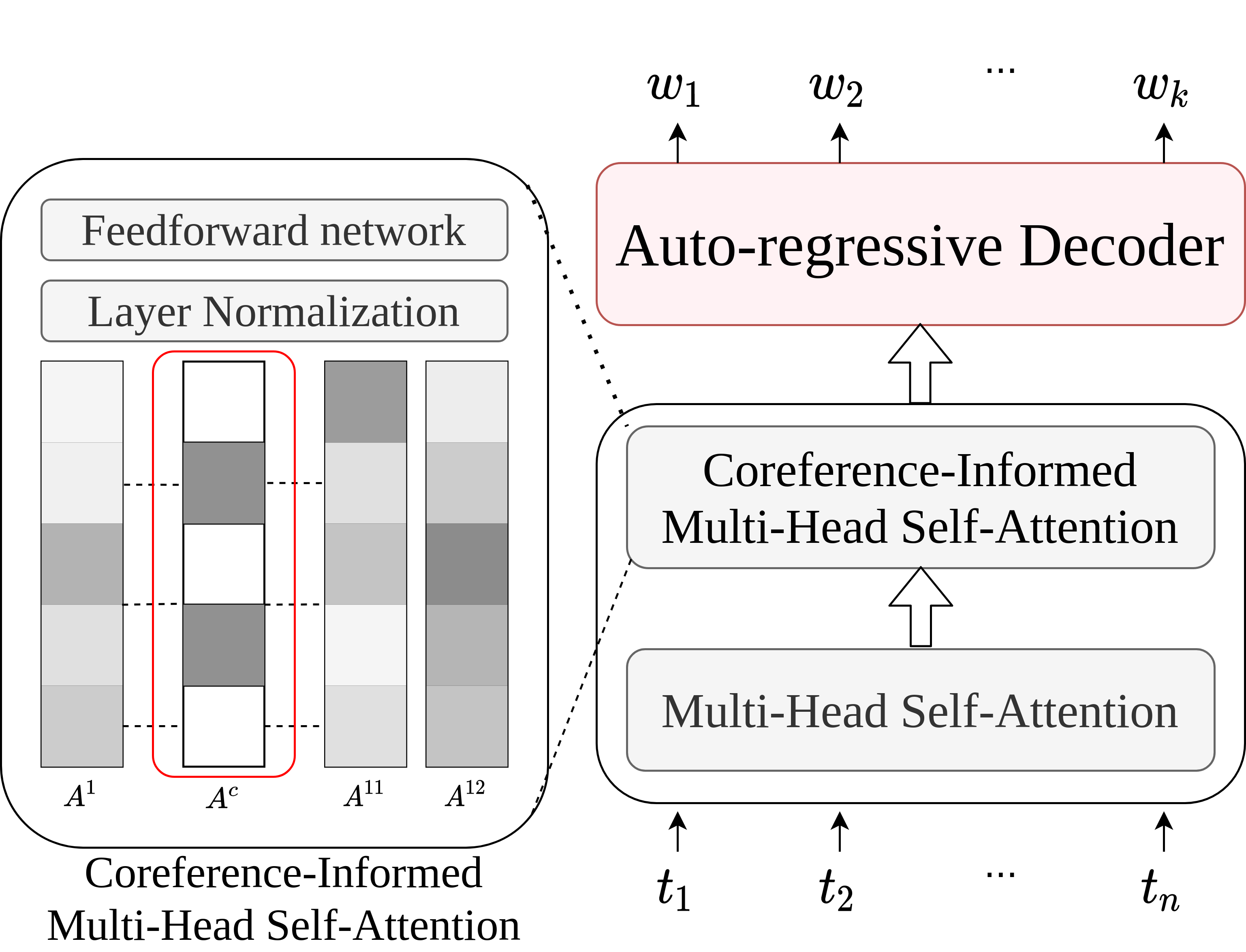}
    \end{center}
    \caption{Architecture overview of the coreference-informed Transformer with attention head manipulation. The second attention head is selected and replaced by a coreference attention weight matrix $A^c$.}
    \label{fig:transformer}
\vspace{-0.2cm}
\end{figure}

While pre-trained models bring significant improvement, they still present insufficient prior knowledge for tasks requiring high-level semantic understanding such as coreference resolution.
In this section, we explore another parameter-free method by directly enhancing the language backbone. Since the encoder of our neural architecture uses the self-attention mechanism, we proposed feature injection by attention weight manipulation. In our case, the encoder of BART \cite{lewis-etal-2020-bart} comprises  6 multi-head self-attention layers, and each layer has 12 heads. To incorporate coreference information, we selected heads and modified them with weights that present coreference mentions (see Figure \ref{fig:transformer}).

\begin{figure}
    \centering
    \includegraphics[width=0.33\textwidth]{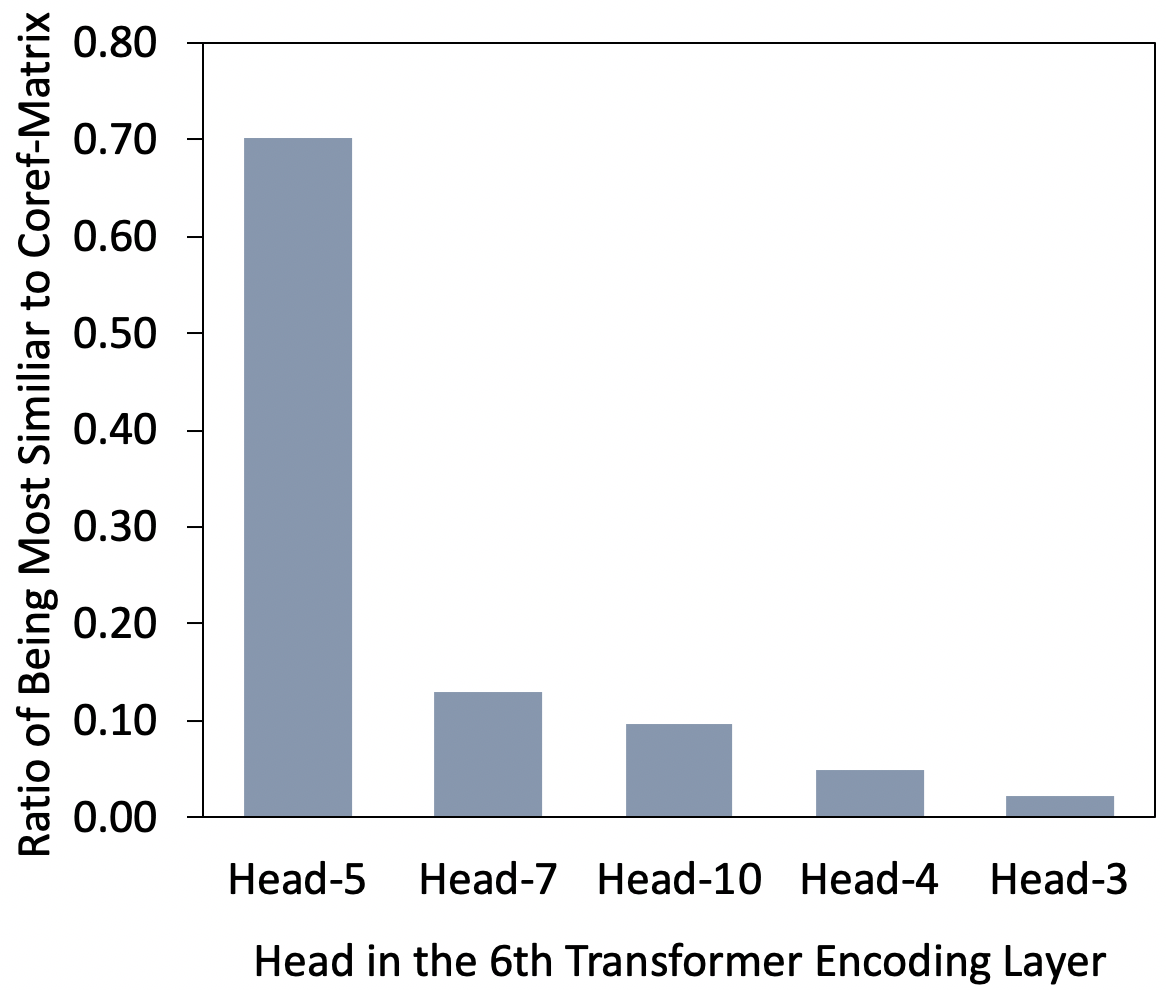}
    \caption{Similarity distribution of head probing with pre-defined coreference matrix. The X-axis shows the heads in the 6-th layer of the Transformer encoder. Values on the Y-axis denote the ratio that a head has the highest similarity with the coreference attention matrix.}
    \label{fig:6_layer_head}
\vspace{-0.2cm}
\end{figure}

\subsubsection{Attention Head Probing and Selection}
\label{probe}
To retain prior knowledge provided by the language backbone as much as possible, we first conduct a probing task to strategically select attention heads.
Since different layers and heads convey linguistic features of different granularity \citep{hewitt-manning-2019-structural}, our target is to find the head that represents the most coreference information.
We probe the attention heads by measuring the cosine similarity between their attention weight matrix $A^o$ and a pre-defined coreference attention matrix $A^c$ as described in Section \ref{attention}:

\vspace{-0.2cm}
\begin{equation}
    head_{probe} = \mathop{\arg\max}_{i}(\cos (A_i^o, A^c))
\end{equation}
\noindent where $A_i^o$ is the attention weight matrix of the original $i$-th head, and $i \in (1,...,N_h)$, $N_h$ is the number of heads in each layer. With all samples in the validation set, we conducted probing on all heads in the 5-th layer and 6-th layer of the \textit{`BART-Base'} encoder. We observed that: (1) in the 5-th layer, the 7-th head obtained the highest similarity score on 95.2\% evaluation samples; (2) in the 6-th layer, the 5-th head obtained the highest similarity score on 68.9\% evaluation samples. The statistics of heads in 6-th encoding layer are shown in Figure \ref{fig:6_layer_head}. 

\subsubsection{Coreference-Informed Multi-Head Self-Attention}
In order to explicitly utilize the coreference information, we replaced the two predominant attention heads with coreference-informed attention weights. The multi-head self-attention layers \cite{vaswani2017attention} are formulated as:
\vspace{-0.0cm}
\begin{equation}
\small
    \mathrm{Attention}(Q,K,V) = \mathrm{Softmax}(\frac{QK^T}{\sqrt{d_k}})V
\end{equation}
\vspace{-0.3cm}
\begin{equation}
\small
    head_i = \mathrm{Attention}(QW_i^Q,KW_i^K,VW_i^V)
\end{equation}
\vspace{-0.4cm}
\begin{equation}
\small
    \mathrm{MHA}(Q,K,V) = \mathrm{Concat}(head_1,...,head_{N_h})
\end{equation}
\vspace{-0.4cm}
\begin{equation}
\small
    \mathrm{FFN}(x_i^l) = \mathrm{ReLU}(x_i^lW_1^F + b_1^F)W_2^F + b_2^F
\end{equation}

\noindent where $Q$, $K$ and $V$ are the sets of queries, keys and values respectively. $W_i$ and $b_i$ are the trainable parameter matrix and bias. $d_k$ is the dimension of keys, $x_i^l$ is the representation of $i$-th token after the $l$-th multi-head self-attention layer. FFN is the point-wise feed forward layer.
Based on the probing analysis in Section \ref{probe}, we selected the 7-th head of 5-th encoding layer, and the 5-th head of 6-th encoding layer for coreference injection, and observed that models with probing selection outperformed that of random head selection.

\begin{table}
\linespread{1.2}
    \centering
    \small
    \begin{tabular}{l|c|c|c|c}
    \hline
         & \textit{\textbf{\# Conv}} & \textit{\textbf{\# Sp}} & \textit{\textbf{\# Turns}} &  \textit{\textbf{\# Ref Len}}\\
    \hline
         Train & 14732 & 2.40 & 11.17 & 23.44 \\
         Validation & 818 & 2.39 & 10.83 & 23.42 \\
         Test & 819 & 2.36 & 11.25 & 23.12 \\
    \hline
    \end{tabular}
    \caption{Data details of the SAMSum corpus. \textit{\# Conv}, \textit{\# Sp}, \textit{\# Turns} and \textit{\# Ref Len} refer to the average number of conversations, speakers, dialogue turns and the average number of words in the gold reference summaries.}
    \label{tab:dataset}
\vspace{-0.2cm}
\end{table}

\begin{table*}[ht]
\linespread{1.25}
\centering
\small
\begin{tabular}{p{4.3cm}|ccc|ccc|ccc}
\hline
\multirow{2}*{\textbf{Model}} & \multicolumn{3}{c}{\textbf{ROUGE-1}} & \multicolumn{3}{c}{\textbf{ROUGE-2}} & \multicolumn{3}{c}{\textbf{ROUGE-L}} \\ 
\cline{2-10}
~ & \textbf{F} &\textbf{P} & \textbf{R} & \textbf{F} &\textbf{P} & \textbf{R} & \textbf{F} &\textbf{P} & \textbf{R}\\
\hline
\textit{Pointer-Generator}* & 40.1 & - & - & 15.3 & - & - & 36.6 & - & - \\
\textit{Fast-Abs-RL-Enhanced}* & 42.0 & - & - & 18.1 & - & - & 39.2 & - & - \\
\textit{DynamicConv-News}* & 45.4 & - & - & 20.6 & - & - & 41.5 & - & - \\
\textit{BART-Large}* & 48.2 & 49.3 & 51.7 & 24.5 & 25.1 & 26.4 & 46.6 & 47.5 & 49.5 \\
 \hline
\textit{Multi-View BART-Large}* & 49.3 & 51.1 & 52.2 & \textbf{25.6} & 26.5 &  \textbf{27.4} & \textbf{47.7} & 49.3 & \textbf{49.9} \\
 \hline
 \hline
\textit{BART-Base} & 48.7 & 50.8 & 51.5 & 23.9 & 25.8 & 24.9 & 45.3 & 48.4 & 47.3 \\
 \hline
\textit{Coref-GNN} & 50.3 & \textbf{56.1} & 50.3 & 24.5 & 27.3 & 24.6 & 46.0 & \textbf{50.9} & 46.8 \\
\textit{Coref-Attention} & \textbf{50.9} & 54.6 & \textbf{52.8} & 25.5 & 27.4 & 26.8 & 46.6 & 50.0 & 48.4 \\
\textit{Coref-Transformer} & 50.3 & 55.5 & 50.9 & 25.1 & \textbf{27.7} & 25.6 & 46.2 & \textbf{50.9} & 46.9 \\
 \hline

\end{tabular}
\caption{ROUGE scores of baselines and proposed models. * denotes the results from \citet{chen-yang-2020-multi}. F, P, and R denote F1 Score, Precision and Recall, respectively.}
\label{tab:results}
\end{table*}

\begin{table}[t!]
\linespread{1.2}
\centering
\small
\resizebox{\linewidth}{!}
{
\begin{tabular}{p{3.0cm}p{0.8cm}<{\centering}p{0.8cm}<{\centering}p{0.8cm}<{\centering}}
\hline
      \textbf{Model}         & \textbf{R-1} & \textbf{R-2} & \textbf{R-L}  \\
\hline
\multicolumn{3}{l}{Models trained with BART-Large} \\
\textit{MV-BART-Large}   &  53.42 &  27.98 &  49.97  \\
\textit{LM-Annotator ($\mathcal{D}_\mathrm{All}$)}   &  53.70   &  28.79 &  50.81  \\
\textit{Our Model (Large)}    &     53.91   &   28.58  &  50.39 \\
\hline
\end{tabular}
}
\caption{\label{table-result-large}ROUGE F1 scores of baselines and our proposed framework. The reported results use the same ROUGE calculation following \cite{feng2021survey} for the benchmarked comparison.}
\end{table}

\section{Experiments}

\subsection{Dataset}
We evaluated the proposed methods on SAMSum \citep{gliwa-etal-2019-samsum}, a dialogue summarization dataset consisting of 16,369 conversations with human-written summaries. Dataset statistics are listed in Table \ref{tab:dataset}.

\subsection{Model Settings}
The vanilla sequence-to-sequence Transformer \cite{vaswani2017attention} was applied as the base architecture. We used the pre-trained \textit{`BART-Base'} \cite{lewis-etal-2020-bart} as language backbone. Then, we enhanced the base model with following three methods:
\noindent\textbf{Coref-GNN}: Incorporating coreference information by the GNN-based fusion (see Section \ref{GNN});
\noindent\textbf{Coref-Attention}: Encoding coreference information by an additional attention layer (see Section \ref{attention});
\noindent\textbf{Coref-Transformer}: Modeling coreference information by the attentive head probing and replacement (see Section \ref{transformer}).
Several baselines were selected for comparison: (1) \textit{Pointer-Generator Network} \citep{see2017get}; (2) \textit{DynamicConv-News} \citep{wu2018pay}; (3) \textit{Fast-Abs-RL-Enhanced} \citep{chen-bansal-2018-fast}; (4) \textit{Multi-View BART} \cite{chen-yang-2020-multi}, which provides the state-of-the-art result.

\subsection{Training Configuration}

The proposed models were implemented in PyTorch \citep{paszke2019pytorch}, and Hugging Face Transformers \citep{wolf-etal-2020-transformers}. The Deep Graph Library (DGL) \citep{wang2019dgl} was used for implementing the \textit{Coref-GNN}. The trainable parameters were optimized by Adam \citep{kingma2014adam}. The learning rate of the GCN component was 1e-3, and that of BART was set at 2e-5. We trained each model for 20 epochs and selected the best checkpoints on the validation set with ROUGE-2 score. All experiments were run on a single Tesla V100 GPU with 16GB memory.

\section{Results}

\subsection{Automatic Evaluation}
We quantitatively evaluated the proposed methods with the standard metric ROUGE \citep{lin-och-2004-automatic}, and reported ROUGE-1, ROUGE-2 and ROUGE-L.\footnote{We used integrated functions in HuggingFace Transformers \cite{wolf-etal-2020-transformers} to calculate ROUGE scores. Note that different libraries may result in different ROUGE scores.} As shown in Table \ref{tab:results}, our base model \textit{BART-Base} outperformed \textit{Fast-Abs-RL-Enhanced} and \textit{DynamicConv-News} significantly, showing the effectiveness of fine-tuning pre-trained language backbones for abstractive dialogue summarization. Adopting \textit{BART-Large} could bring about relative 5\% improvement, while it doubled the parameter size and training time of \textit{BART-Base}. As shown in Table \ref{tab:results}, compared with the base model \textit{BART-Base}, the performance is improved significantly by our proposed methods. In particular, \textit{Coref-Attention} performed best with 4.95\%, 6.69\% and 2.87\% relative F-measure score improvement, and \textit{Coref-GNN} achieved the highest scores on precision with 10.43\% on ROUGE-1, 5.81\% on ROUGE-2 and 5.17\% on ROUGE-L. \textit{Coref-Transformer} also showed consistent improvement.

Moreover, compared with the \textit{BART-Base} model \cite{lewis-etal-2020-bart}, the proposed coref-models performed better on ROUGE-1 scores, especially on the precision metrics. More specifically, precision scores are improved 9.78\%, 6.85\%, and 8.61\% relatively by \textit{Coref-GNN}, \textit{Coref-Attention} and \textit{Coref-Transformer}, respectively. For ROUGE-2 and ROUGE-L, our models also obtain comparable performance.
Recently, \citet{feng2021survey} conducted a benchmarked comparison of state-of-the-art dialogue summarizers. As shown in Table \ref{table-result-large}, our method (trained with \textit{BART-Large}) is comparable to \textit{MV-BART-Large} \cite{chen-yang-2020-multi} and \textit{LM-Annotator ($\mathcal{D}_\textit{All}$)} \cite{feng2021survey}.

\begin{table}[t!]
\linespread{1.2}
    \centering
    \small
    \begin{tabular}{p{2.9cm}l}
    \hline
         \textbf{Model} & \textbf{Average \# Words}  \\
         \hline
         Reference & 23.12 $\pm$ 12.20 \\
         \textit{BART-Base} & 22.72 $\pm$ 10.78\\
         \hline
         \textit{Coref-GNN} & 19.62 $\pm$ 8.75 \\
         \textit{Coref-Attention} & 21.68 $\pm$ 10.27 \\
         \textit{Coref-Transformer} & 20.54 $\pm$ 9.39\\
         \hline
    \end{tabular}
    \caption{Average word number with standard deviations of generated summaries.}
    \label{tab:length}
\end{table}

\begin{table}[t!]
\linespread{1.2}
    \centering
    \small
    \begin{tabular}{p{3cm}c}
    \hline
         \textbf{Model} & \textbf{Average Scores}  \\
         \hline
         \textit{BART-Base} & 0.60\\
         \hline
         \textit{Coref-GNN} & 0.84 \\
         \textit{Coref-Attention} &  \textbf{1.16} \\
         \textit{Coref-Transformer} &  0.96\\
         \hline
    \end{tabular}
    \caption{Human evaluation results: each summary is scored on the scale of [-2, 0, 2] as \citep{chen-yang-2020-multi}. Reported scores are averaged on 100 samples.}
    \label{tab:HumanEval}
\vspace{-0.2cm}
\end{table}

\begin{table*}[ht]
\linespread{1.2}
    \centering
    \small
    \begin{tabular}{l|c|c|c|c}
    \hline
         \textbf{Model} & \textbf{Missing Information} & \textbf{Redundant Information} & \textbf{Wrong Reference} & \textbf{Incorrect Reasoning}\\
         \hline
         \textit{Base Model} & 34 & 26 & 22 & 20 \\
         \hline
         \textit{Coref-GNN} & 32 [5.8\% $\downarrow$] & 8 [69\% $\downarrow$] & 14 [36\% $\downarrow$] & 16 [20\% $\downarrow$] \\
         \textit{Coref-Attention} & \textbf{28} [17\% $\downarrow$] & \textbf{4} [84\% $\downarrow$] & \textbf{12} [45\% $\downarrow$] & \textbf{9} [55\% $\downarrow$] \\
         \textit{Coref-Transformer} & 32 [5.8\% $\downarrow$] & 12 [53\% $\downarrow$] & 14 [36\% $\downarrow$] & 12 [40\% $\downarrow$] \\
         \hline
    \end{tabular}
    \caption{Percentage of typical errors in summaries generated by the baseline and our proposed models. Values in brackets denote the relative decrease compared with the base model.} 
    \label{tab:errors}
\end{table*}

\begin{table*}[ht!]
\linespread{1.1}
    \centering
    \small
    \begin{tabular}{m{.54\textwidth}| m{.2\textwidth}| m{.2\textwidth}}
    \hline
         \textbf{Conversation (abbreviated)} & \textit{\textbf{BART-Base}} & \textit{\textbf{Coref-Attention}}\\
         \hline
          (\romannumeral1)  ... {\color{teal} Ivan} : so youre coming to the wedding {\color{blue} Eric}: your brother's {\color{teal} Ivan}: yea {\color{blue} Eric}: i dont know mannn {\color{teal} Ivan}: YOU DONT KNOW??  {\color{blue} Eric}: i just have a lot to do at home, plus i dont know if my parents would let me {\color{teal} Ivan}: ill take care of your parents {\color{blue} Eric}: youre telling me you have the guts to talk to them XD {\color{teal} Ivan}: thats my problem {\color{blue} Eric}: okay man, if you say so {\color{teal} Ivan}: yea just be there {\color{blue} Eric}: alright 
          & Eric is not sure if he's going to the wedding, because he has a lot to do at home and doesn't know if his parents would let him. {\color{red} Ivan will come to Eric's wedding}.
          & Eric is coming to Ivan's brother's wedding. Eric has a lot to do at home and he can't take care of his parents. Ivan will be there.\\
         \hline
         
          (\romannumeral2) {\color{teal} Derek McCarthy}: Filip - are you around? Would you have an Android cable I could borrow for an hour?  ...  {\color{blue} Tommy} : I am in Poland but can ring my wife and she will give you one ... {\color{blue} Tommy}: 67 glenoaks close {\color{teal} Derek McCarthy}: That would be great if you could!! ... {\color{blue} Tommy}: Sent her msg. She will give it to you. Approx time when she will be at home is 8:15 pm {\color{teal} Derek McCarthy}: Thanks again!! ... 
          & {\color{red}Tommy} will call his wife to {\color{red}borrow} a phone charger {\color{red}from} Derek McCarthy. {\color{red}Tommy} will be at home at 8:15 pm.
          & Filip will lend Derek McCarthy his Android cable. He will call his wife at 67 glenoaks close.\\
          \hline
          \hline
          (\romannumeral3) {\color{teal}Ann}: Congratulations!! {\color{teal}Ann}: You did great, both of you! {\color{orange}Sue}: Thanks, Ann {\color{blue}Julie}: I'm glad it's over! {\color{blue}Julie}: That's co cute of you, my girl! {\color{teal}Ann}: Let's have a little celebration tonight! {\color{orange}Sue}: I'm in {\color{blue}Julie}: me too!!! aww
          & Ann congratulates Sue and Julie on their success. Ann and Julie will celebrate tonight. 
          & {\color{red} Ann and Julie} are congratulating {\color{red} Sue} on their success.\\
          \hline
    \end{tabular}
    \caption{Three examples of generated summaries: For conversation \textit{{\romannumeral 1}}  and conversation \textit{{\romannumeral2}}, \textit{Coref-Attention} model generated correct summaries by incorporating coreference information. \textit{Coref-Attention} model generated an imperfect summary for conversation \textit{{\romannumeral 3}} due to inaccurate coreference resolution provided.}
    \label{tab:sample_analysis}
\vspace{-0.2cm}
\end{table*}

As shown in Table 2, we also observed that the most significant improvement is on the precision scores while the recall scores remains comparable with strong baselines. Moreover, as shown in Table \ref{tab:length}, the average length of generated summaries of the base model is 22.72, and that of the \textit{coref-}models is slightly shorter. We speculated that the proposed models tend to generate more concise summaries while preserving the important information, which is also supported by the analysis in Section \ref{ssec:qualitative_analysis}.

\subsection{Human Evaluation}
As the example shown in Figure \ref{fig:example}, ROUGE scores are insensitive to semantic errors such as incorrect reference, thus we conducted human evaluation to complement objective metrics. Following \citet{gliwa-etal-2019-samsum} and \citet{chen-yang-2020-multi}, each summary is scored on the scale of [-2, 0, 2], where -2 means the summary is unacceptable with the wrong reference, extracted irrelevant information or does not make logical sense, 0 means the summary is acceptable but lacks of important information converge, and 2 refers to a good summary which is concise and informative. We randomly selected 100 test samples, and scored the summaries generated by the base model, \textit{Coref-GNN}, \textit{Coref-Attention} and \textit{Coref-Transformer}. Four linguistic experts conducted the human evaluation, and their average scores are reported in Table \ref{tab:HumanEval}. Compared with the base model, our \textit{coref-}models obtain higher scores in human ratings, which is consistent with the quantitative ROUGE results.

\section{Analysis}
\subsection{Quantitative Analysis}
\label{ssec:qualitative_analysis}

To further evaluate the generation quality and effectiveness of coreference fusion for dialogue summarization, we annotated four types of common errors in the automatic summaries:\\
\noindent \textbf{Missing Information}: The content is incomplete in the generated summary compared with the human-written reference.\\
\noindent \textbf{Redundant Information}: There is redundant content in the generated summary compared with the human-written reference.\\
\noindent \textbf{Wrong References}: The actions are associated with the wrong interlocutors or mentions (\emph{e.g.,} In the example of Figure \ref{fig:example}, the summary generated by base model confused \textit{``Payton''} and \textit{``Max''} in the actions of \textit{``look for good places to buy clothes''} and \textit{``love reading books''}).\\
\noindent \textbf{Incorrect Reasoning}: The model incorrectly reasons the conclusion from context of multiple dialogue turns. Moreover, wrong reference and incorrect reasoning will lead to factual inconsistency from source content. \newline
We randomly sampled 100 conversations in the test set and manually annotated the summaries generated by the base and our proposed models with the four error types.
As shown Table \ref{tab:errors}, 34\% of summaries generated by the base model cannot summarize all the information included in the gold references, and models with coreference fusion improve the information coverage marginally. Coreference-aware models essentially reduced the redundant information: 84\% relative reduction by \textit{Coref-Attention}, 69\% relative reduction by \textit{Coref-GNN}, and 53\% relative reduction by \textit{Coref-Transformer}. \textit{Coref-Attention} model also performed best on reducing 45\% of wrong reference errors relatively, \textit{Coref-GNN} and \textit{Coref-Transformer} both relatively reduced 36\% of that. Encoding coreference information by an additional attention layer substantially improves the reasoning capability by reducing 55\% relatively in incorrect reasoning, \textit{Coref-Transformer} and \textit{Coref-GNN} also relatively reduced this error by 40\% and 20\% compared with the base model.
This shows our models can generate more concise summaries with less redundant content, and incorporating coreference information is helpful to reduce wrong references, and conduct better multi-turn reasoning. 

\subsection{Sample Analysis}
Here we conducted a sample analysis as in \cite{lewis-etal-2020-bart}.
Table \ref{tab:sample_analysis} shows 3 examples along with their corresponding summaries from the \textit{BART-Base} and \textit{Coref-Attention} model.
Conversation \textit{{\romannumeral 1}} and \textit{{\romannumeral 2}} contain multiple interlocutors and referrals. The base model made some referring mistakes:
(1) in conversation \textit{{\romannumeral 1}}, \textit{``your brother's wedding''} should refer to \textit{``Ivan's brother's wedding''}; (2) in conversation \textit{{\romannumeral 2}}, since \textit{``Fillip''} and \textit{``Tommy''} are exactly the same person, pronouns \textit{``you''} and \textit{``I''} in \textit{``Would you have an Android cable I could borrow...''} should refer to \textit{``Tommy''} and \textit{``Derek McCarthy''}, respectively.
In contrast, the \textit{Coref-Attention} model was able to make correct statements. 
However, if the coreference resolution quality is poor, the coreference-aware models will be affected. For example, in the conversation \textit{{\romannumeral 3}}, when the pronouns \textit{``you''} and \textit{``my girl''} in \textit{``Julie: That's co cute of you, my girl''} are wrongly included in the coreference cluster of \textit{``Julie''}, the model will also make referring mistakes in the summary .

\section{Conclusion}
In this paper, we investigated the effectiveness of utilizing coreference information for summarizing multi-party conversations. We proposed three approaches to explicitly incorporate coreference information into neural abstractive dialogue summarization: (1) GNN-based coreference fusion; (2) coreference-guided attention; and (3) coreference-informed Transformer. These methods can be adopted on various neural architectures. Quantitative results and human analysis suggest that coreference information helps track referring chains in conversations. Our proposed models compare favorably with baselines without coreference guidance and generate summaries with higher factual consistency.
Our work provides empirical evidence that coreference is useful in dialogue summarization and opens up new possibilities of exploiting coreference for other dialogue related tasks.

\section*{Acknowledgments}
This research was supported by funding from the Institute for Infocomm Research (I2R) under A*STAR ARES, Singapore. We thank Ai Ti Aw and Minh Nguyen for insightful discussions. We also thank the anonymous reviewers for their precious feedback to help improve and extend this piece of work.


\bibliographystyle{acl_natbib}
\bibliography{acl2021}


\end{document}